\documentclass{article}

     \usepackage[nonatbib, preprint]{neurips_2020}

\usepackage[utf8]{inputenc} 
\usepackage[T1]{fontenc}    
\usepackage{hyperref}       
\usepackage{url}            
\usepackage{booktabs}       
\usepackage{amsfonts}       
\usepackage{nicefrac}       
\usepackage{microtype}      
\usepackage{graphicx}
\usepackage{float}

\title{Sketching the Future (STF): Applying Conditional Control Techniques to Text-to-Video Models}

\author{
  Rohan Dhesikan \\
  Carnegie Mellon Univeristy \\
  \texttt{rdhesika@andrew.cmu.edu}
  \And
  Vignesh Rajmohan \\
  Carnegie Mellon University \\
  \texttt{vrajmon15@gmail.com} \\
  \texttt{vrajmoha@andrew.cmu.edu}
}

\begin{document}

\maketitle

\begin{abstract}
The proliferation of video content demands efficient and flexible neural network-based approaches for generating new video content. In this paper, we propose a novel approach that combines zero-shot text-to-video generation with ControlNet \cite{controlnet} to improve the output of these models. Our method takes multiple sketched frames as input and generates video output that matches the flow of these frames, building upon the Text-to-Video Zero architecture \cite{text_to_video_zero} and incorporating ControlNet to enable additional input conditions. By first interpolating frames between the inputted sketches and then running Text-to-Video Zero using the new interpolated frames video as the control technique, we leverage the benefits of both zero-shot text-to-video generation and the robust control provided by ControlNet. Experiments demonstrate that our method excels at producing high-quality and remarkably consistent video content that more accurately aligns with the user's intended motion for the subject within the video. We provide a comprehensive resource package, including a demo video, project website, open-source GitHub repository, and a Colab playground to foster further research and application of our proposed method.
\end{abstract}

\section{Introduction}

The rapid growth of video content on the internet has led to a surge of interest in neural network-based approaches for generating new video content. However, training Text-to-Video models is challenging due to the lack of open datasets of labeled video data. Furthermore, generating video from existing Text-to-Video models is difficult due to the nature of prompts. To address these challenges, we propose a novel approach that combines the benefits of zero-shot text-to-video generation with the robust control provided by ControlNet \cite{controlnet}.

Our model is built upon the Text-to-Video Zero architecture \cite{text_to_video_zero}, which allows for low-cost video generation by leveraging the power of existing text-to-image synthesis methods, such as Stable Diffusion \cite{stable_diffusion}. The key modifications we introduce include enriching the latent codes of the generated frames with motion dynamics and reprogramming frame-level self-attention using a new cross-frame attention mechanism. These modifications ensure global scene and background time consistency, as well as the preservation of the context, appearance, and identity of the foreground object.

To further enhance the control over the generated video content, we incorporate the ControlNet structure. ControlNet enables additional input conditions, such as edge maps, segmentation maps, and keypoints, and can be trained on a small dataset in an end-to-end manner. By combining Text-to-Video Zero with ControlNet, we create a powerful and flexible framework for generating and controlling video content using minimal resources.

Our model takes multiple sketched frames as input and generates video output that matches the flow of these frames. We first interpolate frames between the inputted sketches and then run Text-to-Video Zero using the new interpolated frames video as the control technique. Experiments show that our method is capable of generating high-quality and remarkably consistent video content with low overhead, despite not being trained on additional video data.

Our approach is not limited to text-to-video synthesis but is also applicable to other tasks, such as conditional and content-specialized video generation, and Video Instruct-Pix2Pix, i.e., instruction-guided video editing. In summary, we present a novel approach that combines the strengths of Text-to-Video Zero \cite{text_to_video_zero} and ControlNet \cite{controlnet}, providing a powerful and flexible framework for generating and controlling video content using minimal resources. This work opens up new possibilities for efficient and effective video generation, catering to various application domains.

\begin{figure}[H]
  \centering
  \includegraphics[width=10cm]{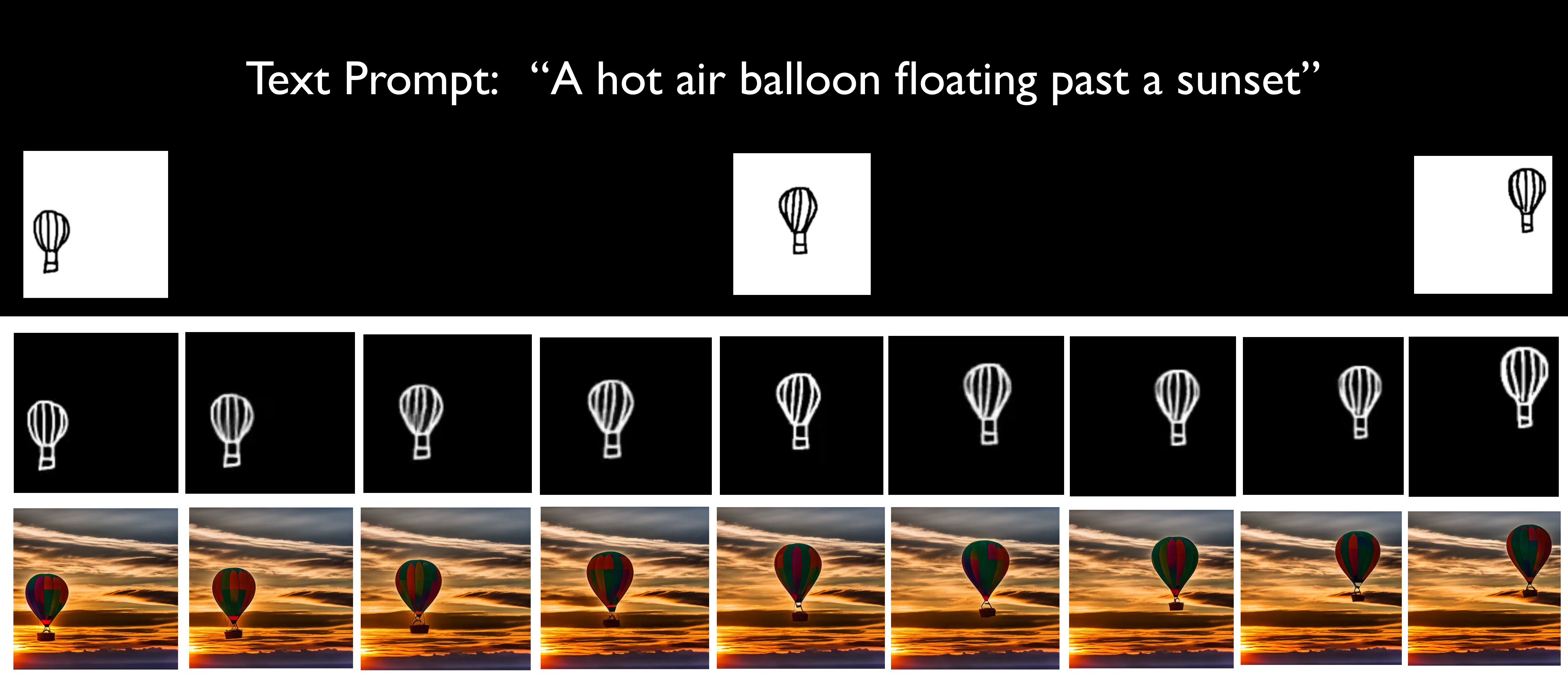}
  \caption{Functional overview of STF: [Input] The basic text prompt on top, [Input] Sketches in the second row, Interpolation in the third
row, and [Output] Resulting frames of the generated video in the bottom row}
\end{figure}

\section{Related Work}
In this section, we discuss two key papers that have significantly influenced our work, namely the Text-to-Video Zero paper and the ControlNet paper.

\subsection{Text-to-Video Zero}
The Text-to-Video Zero paper \cite{text_to_video_zero} addresses the challenge of generating videos from textual prompts without relying on computationally heavy training and large-scale video datasets. The authors introduce a zero-shot text-to-video generation approach that leverages existing text-to-image synthesis methods, such as Stable Diffusion \cite{stable_diffusion}, adapting them for the video domain. This method includes two key modifications: (i) enriching the latent codes of generated frames with motion dynamics for maintaining global scene and background consistency, and (ii) reprogramming frame-level self-attention with a new cross-frame attention mechanism, which focuses on the first frame to preserve context, appearance, and object identity. Experimental results demonstrate that this approach can generate high-quality and consistent videos with low overhead and is applicable to various tasks, including conditional and content-specialized video generation, and instruction-guided video editing.

\subsection{ControlNet}
The ControlNet paper \cite{controlnet} presents a neural network structure designed to control pretrained large diffusion models by supporting additional input conditions. ControlNet learns task-specific conditions in an end-to-end manner and exhibits robust learning even with small training datasets (< 50k). Training a ControlNet is as fast as fine-tuning a diffusion model, making it possible to train on personal devices or scale to large datasets on powerful computation clusters. The authors demonstrate that large diffusion models, such as Stable Diffusion \cite{stable_diffusion}, can be augmented with ControlNets to enable conditional inputs like edge maps, segmentation maps, and keypoints, which may enrich methods to control large diffusion models and facilitate related applications.

Our work builds upon the Text-to-Video Zero architecture by incorporating ControlNet to improve the output of these models. By combining these techniques, we aim to create a more efficient and flexible approach for generating video content from textual prompts and multiple sketched frames.

\section{Method}
Our proposed method aims to improve the output quality of Text-to-Video models by incorporating conditional control techniques. Specifically, our model takes multiple frame and time pairs as input and generates video output that "matches" those frames and time pairs.

Intuitively, our idea makes sense because generating video from text prompts is a challenging task that often results in unrealistic or incorrect outputs. By allowing the user to provide input in the form of frames and time pairs, our model is better able to understand the desired outcome and generate more accurate and desirable video content.

Our proposed approach builds upon prior work in the area of Text-to-Image Diffusion Models. These models utilize a stable diffusion process to generate images from text prompts. We use a similar approach to create a Stable Diffusion (SD) model and a locked SD model that can be trained to generate video from image-scribble pairs. This is illustrated in the ControlNet paper [2].

\begin{figure}[H]
  \centering
  \includegraphics[width=10cm]{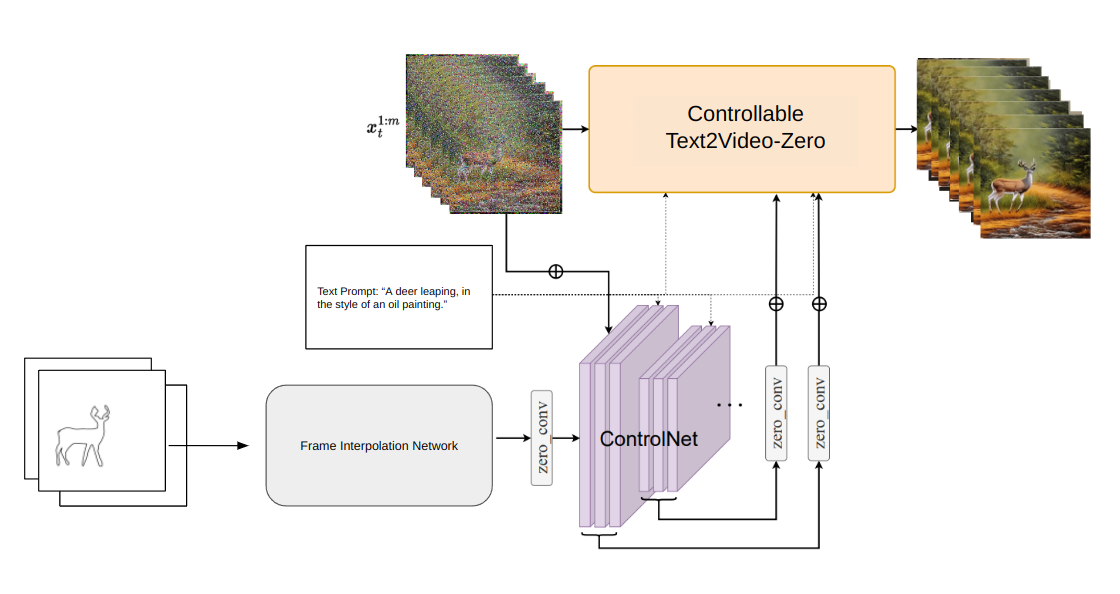}
  \caption{Architecture overview for ControlNet}
\end{figure}

In addition, our work is heavily inspired by the zero-shot video generation approach that Text2Video-Zero showcased [1]. Text2Video-Zero applies Stable Diffusion to generate video from text by applying techniques to keep the foreground and background consistent between frames. 

\begin{figure}[H]
  \centering
  \includegraphics[width=14cm]{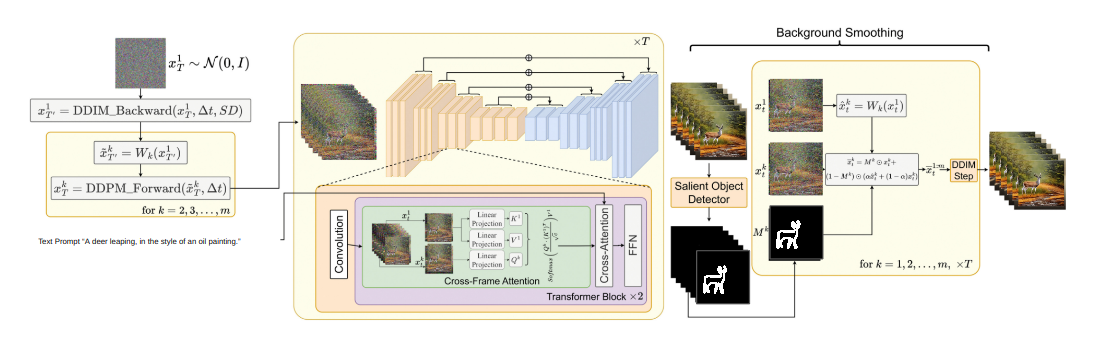}
  \caption{Architecture of Controllable Text2Video}
\end{figure}

We randomly sample latent codes and subsequently use Stable Diffusion to generate modified latent codes. Then, we enhance the latent codes with motion dynamics to simulate the motion in the video. Subsequently, the latent codes are pipelined through the ControlNet model with added cross-frame attention to ensure the foreground object remains consistent. 

To expand more on the cross-frame attention, each self-attention layer combines a feature map into a query, key and feature. As done in the Text2Video-Zero paper, we add a cross-frame attention layer on each frame against the first frame. This maintains the identity of the foreground and background objects.

\begin{figure}[H]
  \centering
  \includegraphics[width=14cm]{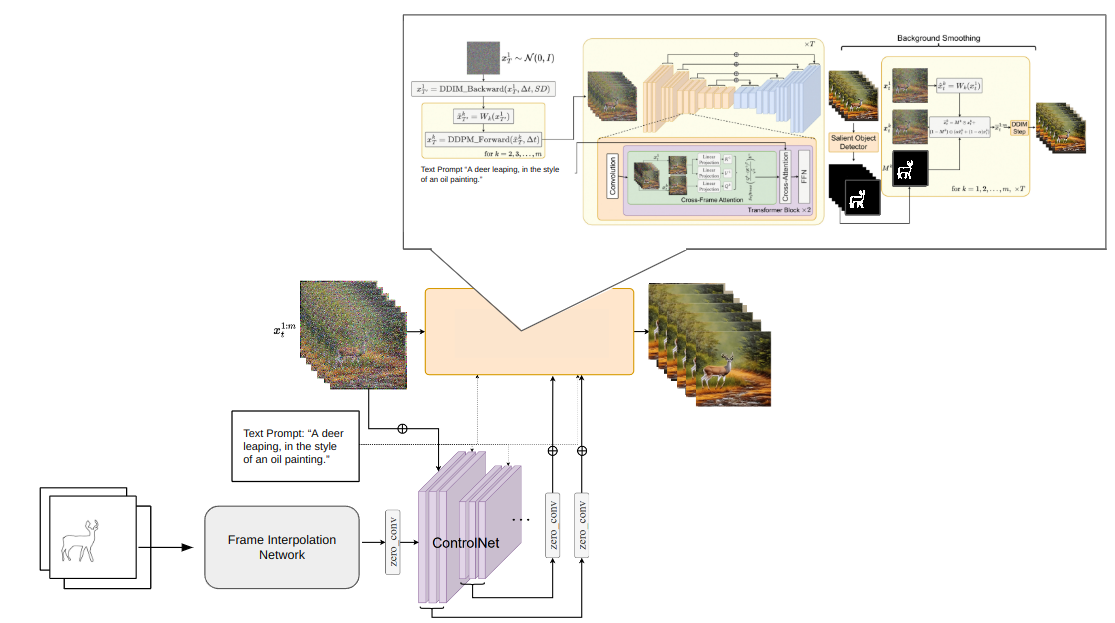}
  \caption{Architecture overview of STF}
  \label{fig:stf_architecture}
\end{figure}

\section{Experiments}

In our experiments, we qualitatively compare the performance of Text2Video-Zero with and without the control provided by our proposed Sketching the Future (STF) approach. Our aim is to demonstrate the effectiveness of STF in guiding the generated video output to match our desired vision, even when the original Text2Video-Zero model fails to capture the intended motion. Our goal with the video models is to generate a video of a man walking from the left side of the frame to the right side. We will look into the effectiveness of Text2Video-Zero with just text prompting, and then STF with sketches for control along with text prompting.

\subsection{Baseline: Text2Video-Zero without Control}

We first test the Text2Video-Zero model on a prompt that describes our desired video: "A man walking on the beach in front of the ocean." The resulting video shows a man's legs walking on a beach towards the camera. Although the beach and ocean context are captured, the video does not convey our vision of the man walking from left to right across the frame.

\begin{figure}[H]
  \centering
  \includegraphics[width=14cm]{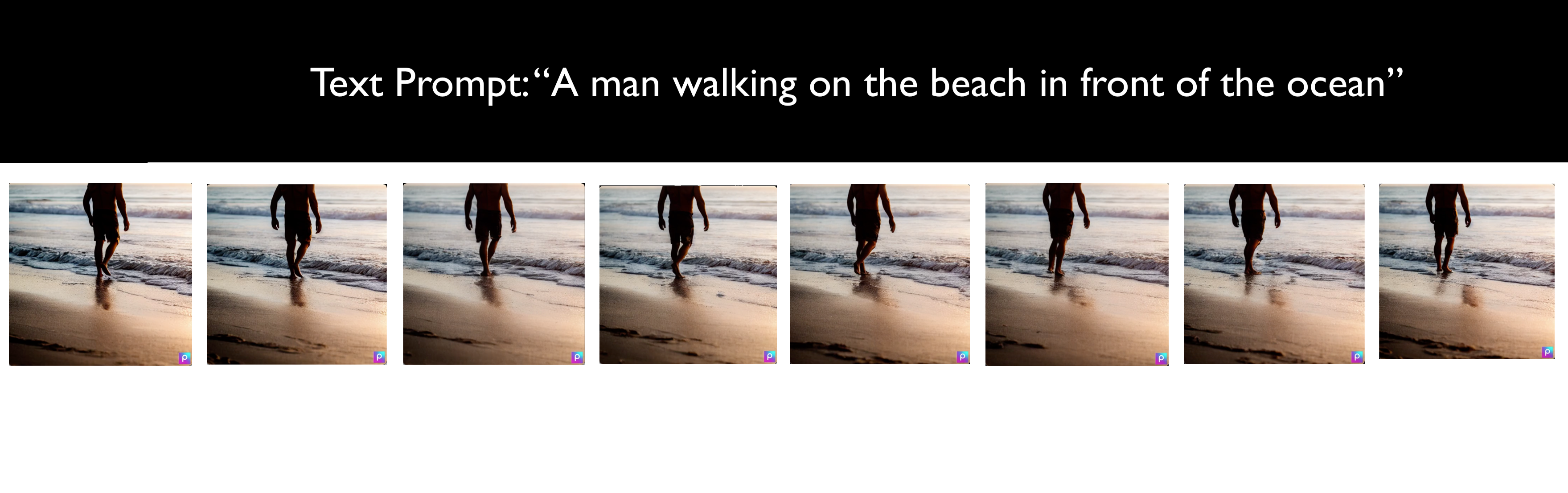}
  \caption{Text2Video-Zero with basic prompt, and resulting frames of the generated video in the bottom row}
  \label{fig:t2v1}
\end{figure}

To better convey our desired motion, we extend the prompt to: "A man walking on the beach in front of the ocean [from left to right]." However, the generated video still does not meet our expectations, as it only shows a close-up of the man's feet, and they remain stationary in the frame without any indication of the "left to right" aspect. This highlights the limitations of the Text2Video-Zero model in capturing specific motion details through text prompts alone.

\begin{figure}[H]
  \centering
  \includegraphics[width=14cm]{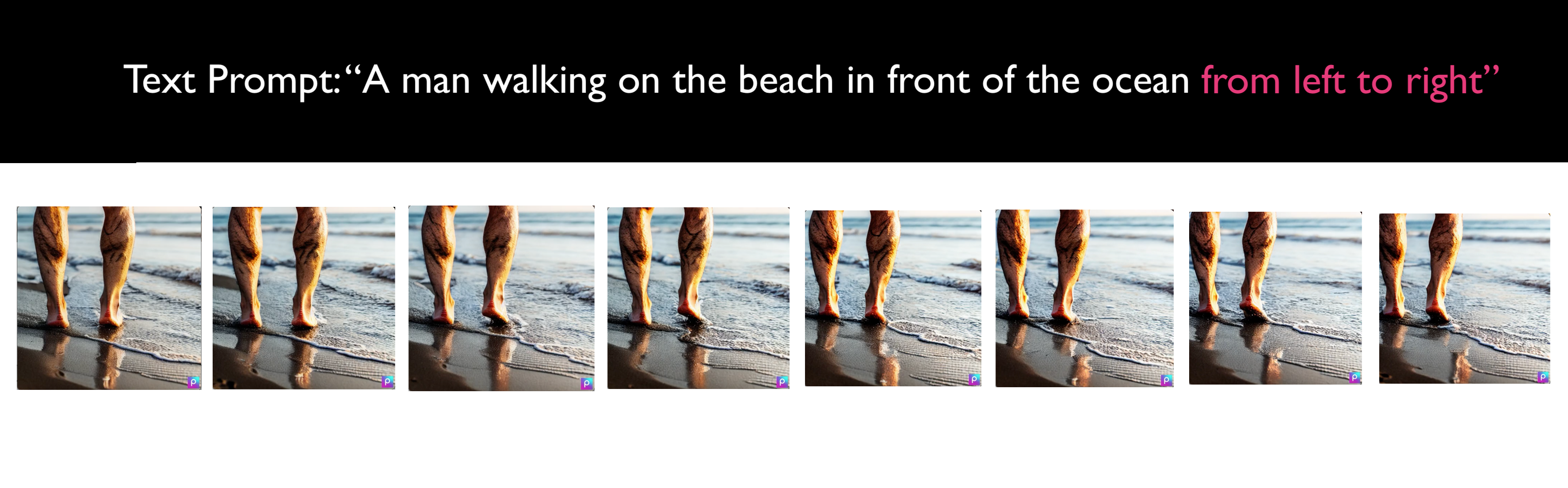}
  \caption{Text2Video-Zero with motion details in prompt, and resulting frames of the generated video in the bottom row}
  \label{fig:t2v2}
\end{figure}

\subsection{Proposed Approach: Sketching the Future}

To address the limitations of Text2Video-Zero, we use our STF approach, which incorporates three sketches of a stick figure man as additional input. The first sketch places the man on the left, the second sketch in the middle, and the third sketch on the right of the frame. These sketches help convey the desired motion more explicitly and provide the model with additional visual cues to generate the intended video.

\begin{figure}[H]
  \centering
  \includegraphics[width=14cm]{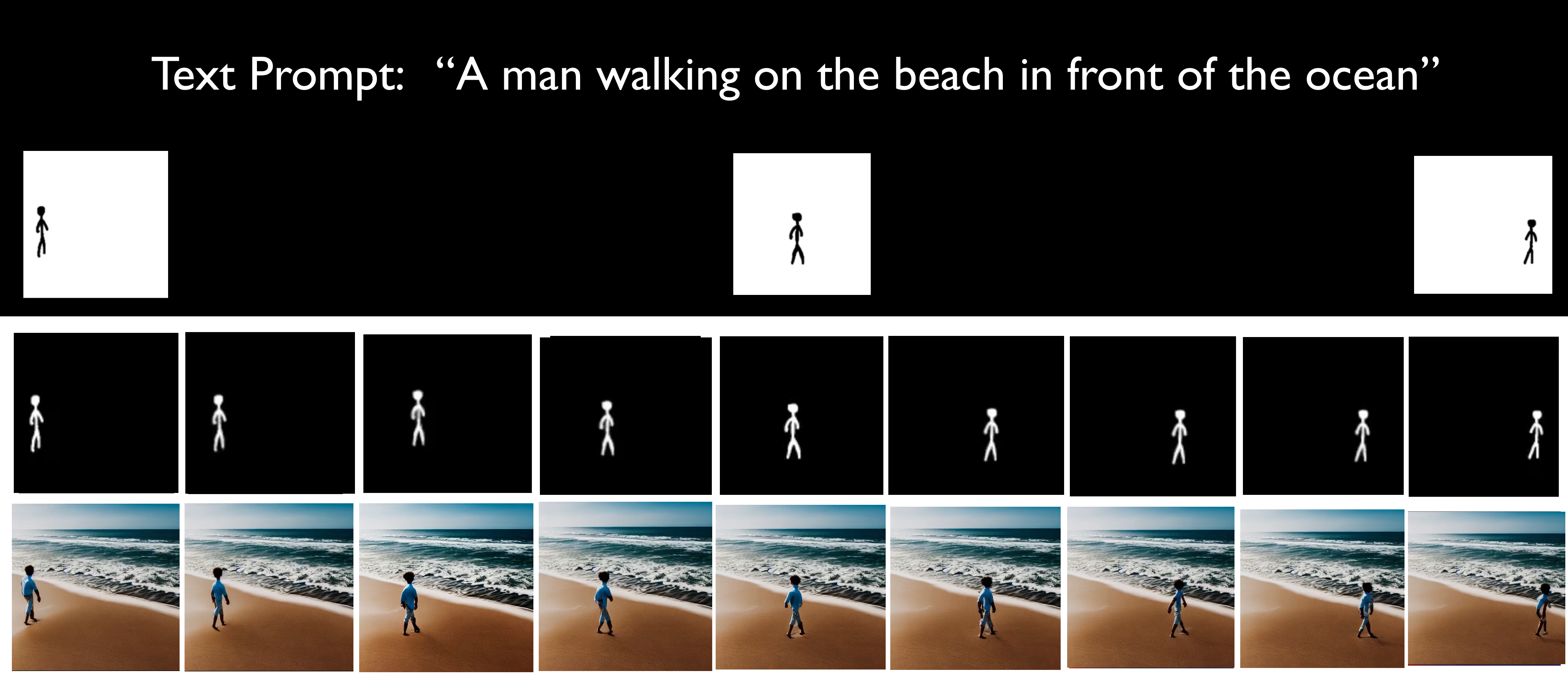}
  \caption{STF with the basic text prompt, and sketches in the second row, interpolation in the third row, and resulting frames of the generated video in the bottom row}
  \label{fig:stf1}
\end{figure}

When we apply STF to the Text2Video-Zero model, the resulting video perfectly matches our vision, with the man walking from the left to the right of the beach. The man's motion is clearly captured, and the background remains consistent throughout the video. This demonstrates the effectiveness of our approach in guiding the video generation process and providing more control over the output, which was not achievable with Text2Video-Zero alone. The success of our STF approach highlights the potential of combining textual prompts with visual input for more accurate and controlled video generation tasks.

\section{Conclusions and Discussions}
In this work, we presented STF (Sketching the Future), a novel approach that combines zero-shot text-to-video generation with ControlNet to improve the output of these models. Our method takes multiple sketched frames as input and generates video output that matches the flow of these frames, providing a more accurate representation of the desired motion compared to Text2Video-Zero with pure prompting.

Our experiments demonstrated that the inclusion of sketched frames allowed STF to generate video content more in line with the desired action specifics. We compared the performance of Text2Video-Zero without control, Text2Video-Zero with additional prompting, and our proposed STF approach. The results showed that STF was capable of generating the desired video content, while the other methods failed to fully capture the intended motion.

The experiments, though qualitative in nature, provided valuable insights into the capabilities of our approach. By focusing on three distinct cases, we were able to showcase the advantages of STF over traditional methods. In the first case, we observed that Text2Video-Zero generated a video that only partially matched the prompt, with the man's legs walking towards the camera rather than from left to right. In the second case, adding an extension to the prompt yielded a video that showed the man's feet but failed to consider the "left to right" aspect of the prompt. Finally, with STF, the generated video accurately depicted the man walking from left to right across the beach, as desired.

These qualitative results highlighted the effectiveness of our approach in generating more accurate video content based on the given prompts and sketches. Our method overcame the limitations of existing text-to-video generation approaches, which often struggle to capture the full semantics of the input prompt. The success of STF, as shown in the experiments, underscores the importance of incorporating additional input conditions, such as sketched frames, to guide the video generation process more effectively.

This work opens up several avenues for future research. One potential direction is to explore the integration of additional types of input modalities, such as audio or text, to provide even more control over the generated video content. Furthermore, it would be interesting to investigate methods for automatically generating the sketched frames from textual descriptions or other forms of input, thereby streamlining the video generation process. Additionally, exploring ways to improve the quality of the generated videos by refining the interpolation process between sketched frames and optimizing the control mechanism could lead to even better results.

Another potential research direction is to evaluate the performance of STF on a wider range of tasks and scenarios, including more complex and dynamic scenes, to further demonstrate its versatility and applicability. This would help establish the method's generalizability and its potential for adoption in various real-world applications.

In conclusion, STF offers a promising approach for generating high-quality and remarkably consistent video content with low overhead, despite not being trained on additional video data. Its ability to leverage multiple sketched frames as input and combine them with powerful techniques like ControlNet makes it a valuable tool for various applications, including conditional and content-specialized video generation, and instruction-guided video editing. By addressing the limitations of existing text-to-video generation methods, STF paves the way for more effective and efficient video content generation in the future.

\section{Open Resources}

In this section, we provide a list of resources related to our STF (Sketching the Future) project. These resources include a demo video, a project website, an open-source GitHub repository, and a Colab playground where users can interactively try out STF.

\begin{enumerate}
\item \textbf{Demo Video}: A video demonstration showcasing the capabilities of our STF approach and the generated video content can be found at the following link: \url{https://www.youtube.com/watch?v=5QH3T7mhomw}.
\item \textbf{Project Website}: For more information about the STF project, including the latest updates, news, and related publications, please visit our project website at: \url{https://sketchingthefuture.github.io/}.

\item \textbf{GitHub Repository}: The source code for STF is openly available on our GitHub repository. Users can access the code, report issues, and contribute to the project at the following link: \url{https://github.com/rohandkn/skribble2vid}.

\begin{figure}[H]
  \centering
  \includegraphics[width=10cm]{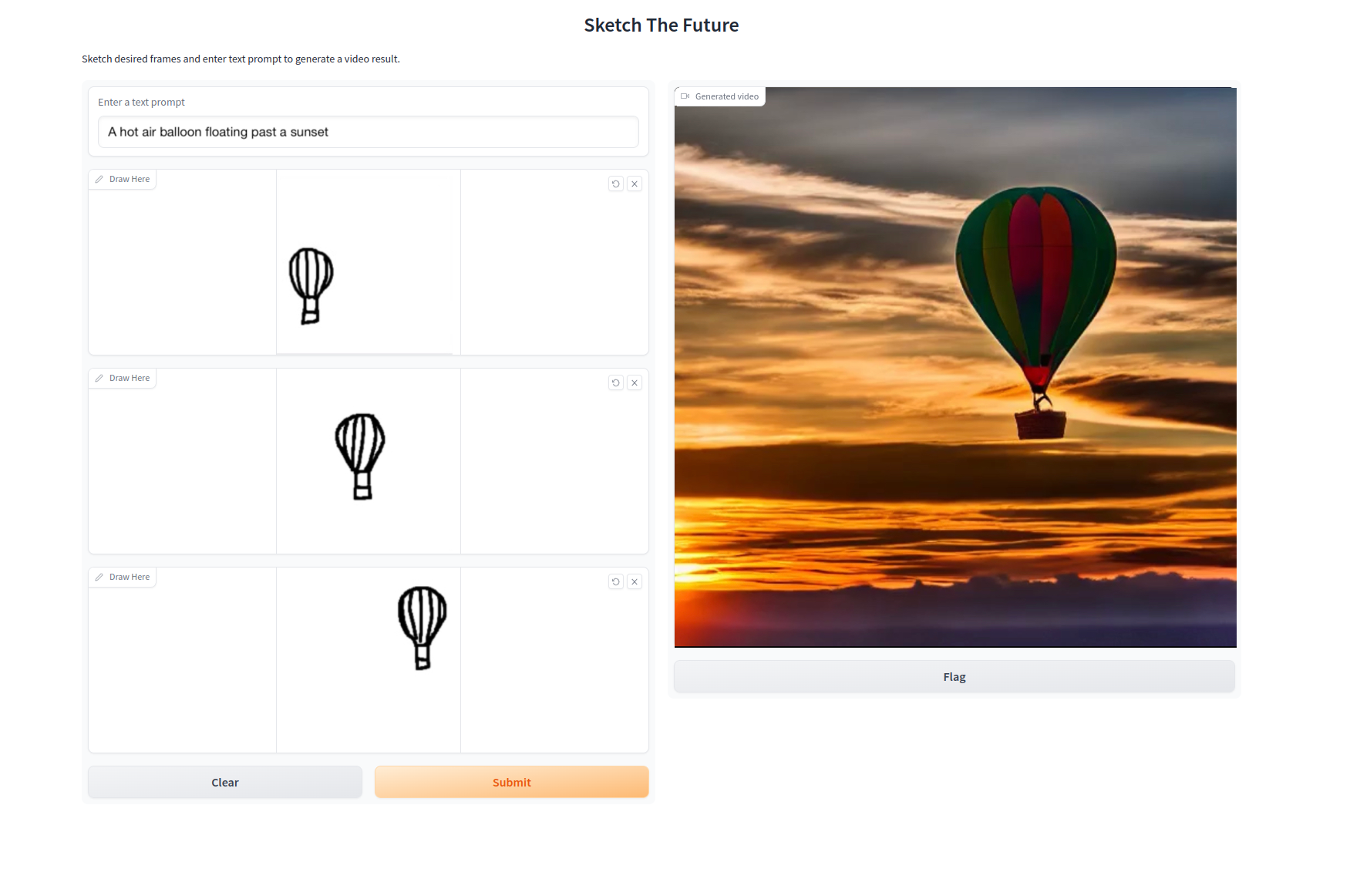}
  \caption{Colab open playground}
  \label{fig:stf1}
\end{figure}

\item \textbf{Colab Playground}: To enable users to interactively try out our STF approach and experiment with different sketched frames and textual prompts, we have created a Colab playground. Users can access the playground at the following link: \url{https://colab.research.google.com/drive/1xc2P3x-10oqsMxA-2SuDlwoE41vCS0m6?usp=sharing}.

\end{enumerate}

We encourage users to explore these resources and provide feedback, which will help us improve the STF project and better understand its potential impact on various industries and applications.

\section*{Broader Impact}

The development of STF (Sketching the Future) has significant implications for a wide range of industries and applications. As a novel approach that combines zero-shot text-to-video generation with ControlNet, STF has the potential to greatly impact the way we generate and consume video content. In this section, we discuss some of the broader impacts of our work, both positive and negative, that may arise from its adoption.

\subsection{Positive Impacts}

\textbf{Creative industries}: STF can be a valuable tool for creative professionals, such as filmmakers, animators, and graphic designers. By enabling the generation of video content from sketched frames and textual prompts, our approach can help streamline the creative process and reduce the time and effort required to create high-quality video content.

\textbf{Advertising and marketing}: The ability to generate custom video content quickly and efficiently can be beneficial for advertising and marketing campaigns. STF can help companies create engaging and targeted promotional materials, enabling them to better reach and resonate with their intended audiences.

\textbf{Education and training}: STF can be used to develop educational materials tailored to specific learning objectives or training requirements. By generating video content that aligns with the desired learning outcomes, our approach can contribute to more effective and engaging educational experiences.

\textbf{Accessibility}: STF has the potential to make video content more accessible for individuals with disabilities. By generating video content that includes captions or other visual aids, our approach can help make information and entertainment more inclusive and accessible to a broader audience.

\subsection{Negative Impacts}

\textbf{Misinformation and deepfakes}: The ability to generate realistic video content from textual prompts and sketched frames raises concerns about the potential for misinformation and deepfake videos. Malicious actors may use STF to create convincing but false video content, which could be used to spread disinformation or manipulate public opinion.

\textbf{Privacy concerns}: The use of STF in surveillance or monitoring applications could potentially infringe on individual privacy. If our approach is used to generate video content that includes recognizable people or places, it may raise ethical and legal concerns related to consent and data protection.

\textbf{Job displacement}: The widespread adoption of STF in industries that rely on manual video content creation could lead to job displacement for some professionals. While our approach can streamline the video creation process, it may also reduce the demand for certain roles in the creative industries, such as animators and video editors.

\subsection{Mitigating Negative Impacts}

To address these potential negative impacts, it is crucial to develop guidelines and regulations around the ethical use of STF and similar technologies. Researchers, developers, and policymakers should work together to establish best practices and standards that prioritize privacy, consent, and the responsible use of video generation technology. Additionally, efforts should be made to educate the public about the capabilities and limitations of STF and other video generation methods, so that they can critically evaluate and interpret the content they encounter.

\section*{Acknowledgements} 
We would like to thank Professor Pathak and the course staff of Visual Learning and Recognition for their support, and Mrinal Verghese for his compute resources. Also we would like to thank ChatGPT for assisting with the writing and organization of this paper.

\end{document}